# Generalized Grasping for Mechanical Grippers for Unknown Objects with Partial Point Cloud Representations

Michael Hegedus[1], Kamal Gupta[1*], and Mehran Mehrandezh[2*]

*Abstract*—We present a generalized grasping algorithm that uses point clouds (i.e. a group of points and their respective surface normals) to discover grasp pose solutions for multiple grasp types, execute by a mechanical gripper, in near real-time. The algorithm introduces two ideas: 1) a histogram of finger contact normals is used to represent a grasp "shape" to guide a gripper orientation search in a histogram of object(s) surface normals, and 2) voxel grid representations of gripper and object(s) are cross-correlated to match finger contact points, i.e. grasp "size", to discover a grasp pose. Constraints, such as collisions with neighbouring objects, are optionally incorporated in the cross-correlation computation. We show via simulations and experiments that 1) grasp poses for three grasp types can be found in near real-time, 2) grasp pose solutions are consistent with respect to voxel resolution changes for both partial and complete point cloud scans, and 3) a planned grasp is executed with a mechanical gripper.

## I. INTRODUCTION

Grasp planning and manipulation is fundamental in a variety of robotic domains, in particular for assisted robotics to aid people to complete a wide variety of tasks. In robotics literature, grasps are planned based on analytical and heuristic approaches to securely grasp an object[1, 2]. For instance, force closure and caging works focus on analytical methods that rely on an object's geometry, kinematic, and/or dynamic equilibrium[3-6]. A limitation for these methods is their design is typically for one specific task wrench (such as resisting gravity) or grasp type (such as a two fingered pinch grasp). As pointed out in [7], even newer learning-based methods are applied using one grasp type, typically a parallel-jaw gripper.

Planning for a single grasp type is rather limiting because the grasp type itself is determined from the task being accomplished, e.g., a power grasp to hold a tool, tripodal to grasp a ball, or precision to hold a pen[7]. Determining a grasp itself is not simply securing an object with parallel jaws, rather it is completing a practical task comprising of multiple subtasks which may require different grasp types. Securing an object is only one sub-task; other task types include object transfer, object manipulation, tool usage, etc. The principle objective of our research is to simultaneously generate grasp poses for different grasp types to accomplish a task.

A feature of our work is discovering grasp candidates without explicitly estimating an object's dynamic equilibrium. We forgo an explicit stability check because our approach is designed to work with partial information about an object; practically, stability can only be guaranteed with complete information about the object. We posit that a grasp taxonomy's grasp type is inherently stable[8, 9]. A finite number of grasp types exist because if applied correctly, they inherently yield a stable grasp. For example, a pinch (or parallel) grasp applied to a plate's corner may not be stable due to torque allowing the plate to rotate between each finger; however, this is a common grasp because more force can be applied at one's fingertips to increase friction and prevent such a rotation. Given rigid objects, applying maximum gripper force can stabilize an object, given the grasp type being attempted inherently yields stability. Discovering a grasp candidate for a grasp type is not 'optimized'; we argue optimizing (or selecting) a grasp should be a final decision because any optimization is dictated by a task. For example, if a task is securing an object, force closure and caging are optimized to restrict an object's dynamic motion, but the consequence is gripper contacts will ideally surround an object. If the task changes to transfer, a suboptimal grasp (in terms of minimizing dynamic motion) is desired to reveal more surface area for another person to grasp.

In grasping literature, there is an implicit theme to place importance on object shape. Object shape requires an appropriate grasp taxonomy[8], but this leads to a significant problem; an infinite number of shapes requires numerous grasp types. A human hand is observed to have over thirty different grasp variations[8-11]. Computing all grasp types for all objects would be an endless task. However, the search space for grasping can be reduced if we focus to generalize grasp types rather than an object's shape. The key idea in our approach is to discover if a grasp type exists or "matches" to a set of contacts on an object's surface; we assume a grasp is achievable if it exists. If a grasp type does not exist, that grasp cannot be executed. Even though over thirty grasp types exist for a human hand, these types can be generalized for a mechanical gripper to complete a task. (i.e. pinching, power, tripodal, parallel, ring, etc.)[11].

Based on the above key idea, we present a generalized grasping approach for mechanical grippers that permits grippers with any number of fingers to discover poses for different grasp types in near real-time. Grasp type matching is performed in a computationally efficient two stage process. First, a set of grasp type orientations that yield a "pure shape" match are discovered by matching histograms between finger contacts' (corresponding to a grasp type) and objects' surface normals. A pure shape match is when a grippers' finger contact normal distribution matches that observed on a set of object surface normals. Second, "size" of grasp type is matched by cross-correlating voxel grids representing the gripper with partially viewed objects. The first stage determines if an observed shape matches a grasp type, and the second stage matches the shape's size/scale to the gripper model. Furthermore, collision constraints, e.g. the gripper palm, can be accommodated in the second stage as one single step by introducing negative penalties during cross-correlation. The second stage also accommodates finger alignment issues; finger contacts may physically match an object surface but gripper contact normals may not perfectly align with observed surface normals. A tolerance rejects poses when the angle between any gripper contact and object surface normal becomes too large.

*Research supported by NSERC Discovery and RTI Grants
[1] School of Engineering Science, Simon Fraser University, Burnaby, BC V5A 1S6, Canada (e-mail: [mhegedus, kamal]@sfu.ca)

[2] Faculty of Engineering and Applied Science, University of Regina, Regina, SK S4S 0A2, Canada (e-mail: mehran.mehrandezh@uregina.ca)

Computational efficiency is achieved by decoupling shape matching (using contact surface normals) from scale (using contact points). A similar matching result may be achieved by cross-correlating gripper contact normals and points in 6-dimensions (6D). However, the computation time to cross-correlate hypercubes in the frequency domain is $O[v^D log(v)]$, where 'v' is the number of voxels needed to create a cube's length, 'D' is the dimensional space, and $v^D$ is total voxels[12]; a decoupled approach in 3D results in significant computational efficiency.

In summary, the original contributions of our work are: 1) introducing a novel approach to represent grasp types and partially scanned objects using surface normal histograms and voxel grids, 2) presenting a data-driven method that discovers different grasp types from a partial scan with no offline-training, 3) integrating collision checks and Cartesian trajectory planning for a grasp pose in a single correlation step, 4) showing this method is scalable for n-contact points and is invariant to point cloud size, and 5) deriving grasp pose solutions for several unique grasp types in near real-time.

II. RELATION TO PREVIOUS WORK

*a) Analytical Grasping Methods*

Analytical grasping approaches tend to apply specific grasp types that use two to four contact points. Some earlier works that present data-driven grasping from a point cloud applied force closure to determine grasp quality[13-15]. These works improved the computation time for force closure analysis by assuming a gripper pre-shape or introduced a heuristic stage that avoids searching for all contact combinations. For example, [13] showed three contact points form a triangular plane that can be decomposed to a series of IF-THEN statements, using contact normals, to determine a force closure grasp. The grasp planner presented in [14] reduced computation time for force closure by adding a pose constraint to align the gripper's palm with the object so all fingers will touch the object simultaneously when closed. The planner presented in [15] reduces computational time by searching for a parallel grasp; the grasp wrench would only be estimated for contacts that geometrically matched a parallel jaw. All of these works demonstrated grasping but these grasps are specialized and represent only one grasp type (or purpose).

Later works generalize force closure to n-contacts, but experiments would demonstrate randomly generated contacts, not considering a gripper's shape and constraints[16-18]. In contrast, our algorithm discovers potential grasp locations for n-contacts while maintaining a gripper's physical constraints.

*b) Geometric Shape Representation*

Spatial features have been extracted from household objects from a single viewpoint to be classified into several different object primitives for grasping[19]. An object primitive simplifies an object's shape and represents it with a-priori known geometric shapes (e.g. cube, cylinder, or sphere)[20, 21]. An object is either represented with a single primitive or can be decomposed into a group of sub-primitives[22-24]. Learning-based methods also explore primitive shapes and discover grasps using simplified shapes[25-27] or similar objects[28, 29]. From these representations, either an analytical or data-driven database method can identify grasp locations. Comparably, object primitives can be viewed as a low resolution, quantized voxel grid setting applied by our algorithm. A key difference is our algorithm does not have a-prior geometric assumptions for an object model; their shape primitive can be thought of as a special case of our algorithm that occurs when using low voxel resolution to represent an object. An object's shape is not the key methodology we apply for grasping—representing the gripper shape and all corresponding grasp types is key.

*c) Machine Learning and Heuristic Approaches*

More recently, machine-learning (ML) based approaches to grasp an object are presented in research. Most ML based grasp planning systems are specific to one grasp type or simply a parallel jaw gripper[1, 2, 7, 29-33]. Recently, [7] demonstrated a learning-based approach able to perform grasps using two grasp types. This algorithm did not select which type is most appropriate but demonstrated their framework can flexibly learn different grasp types. Another method selects between suction and pinch grasp modalities to retrieve objects from a container[34]. Through human demonstration, earlier works demonstrated different grasp types to grasp a single object[35]. Even earlier work demonstrated database driven grasping for approximately 7256 objects associated with 238,737 grasps for several different grippers[36]. These methods require large databases or significant training, which takes time to develop, and similar data-driven approaches do not scale well. For example, adding a new gripper or grasp type would significantly increase a database's size, and the grasp type added may become more difficult to discern from others already embedded.

Our work differs as we demonstrate different grasp types can be discovered in near real-time using conventional signal processing techniques, with no training or large databases. For mechanical grippers, our approach evaluates several grasp types using partial point cloud data and can select a grasp type depending on the task specified.

III. SYSTEM OVERVIEW

Our system determines a grasp pose by cross-correlating a gripper's grasp type shape with an object's surface. Furthermore, by adding negative penalties within the correlation computation, a collision free grasp and a "straight line path" to the object can be predicted in one integrated step. We assume object(s) being modelled are represented by a point cloud, a collection of points located on an object's

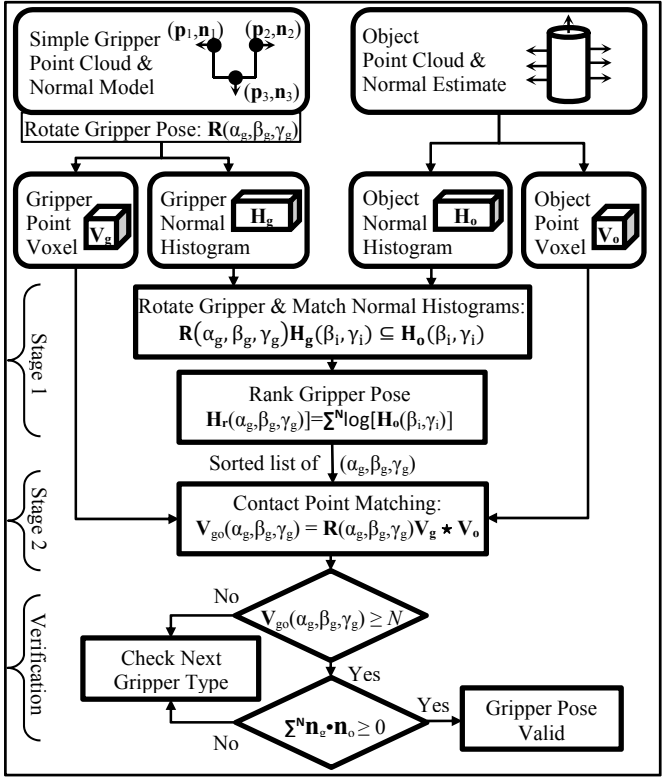

Figure 1. Grasp Planner Pipeline

## a) Stage One to Match Grasp Type Shape

Stage 1 determines if a grasp type shape (not the scale) "matches" any part of viewed unknown objects, i.e. a grasp type's set of contact normals are observed on any scanned object. A straightforward measure for this is matching object surface normal bins to non-zero bin locations within the grasp histogram. A grasp type's contact normals used to create $\mathbf{H_g}$, defined a-priori, are rotated over the gripper's pose space using Euler angles. For each gripper rotation, a rank quantifies how many object surface normals match desired gripper contacts. Heuristically, this step selects and ranks grasp orientations (i.e. top-down, sideways, etc.) that match the grasp type's shape to any observed objects. These orientations will need further investigation to determine if the shape's scale matches the gripper in a subsequent stage. A Stage 1 example, shown in Fig. 2, illustrates a parallel jaw gripper with two contact points matching a partially observed box (in 2D). In this example, the gripper model contact normals (shown as lines), align with the observed object normals only when the gripper pose is rotated ±90°, using rotation matrix $\mathbf{R} \in \mathbb{R}^{3 \times 3}$ to rotate the gripper model's point normals. Green points are inverted gripper contact normals. Blue points are outer surface normals associated to the object.

## b) Stage Two to Match Grasp Type Scale

Stage 2 determines if a grasp type's contacts points "match" a set of observed surface points on the object. Gripper orientations that satisfy Stage 1 are used to rotate a grasp type's point cloud model. Rotated points are then inserted into a gripper voxel grid ($\mathbf{V_g}$) and cross-correlated with the object's voxel grid (i.e. $\mathbf{V_{go}} = \mathbf{RV_g} \star \mathbf{V_o}$), where $\star$ denotes a cross-correlation operation. Peaks from correlation greater than or equal to the number of grasp type contacts identify locations for a potential grasp. Lastly, a final step revisits these physical locations, finds the gripper contact points closest to a surface point in the object point cloud, and verifies each contact normal and object surface normal are in a "similar" direction. If they are, the grasp type's complete pose is inserted into a list for the robot to attempt. Stage 2 for the parallel jaw gripper is shown in Fig. 3, where parallel jaw gripper's green contact points are cross-correlated to an object's blue surface points; from cross-correlation, an orange 'X' identifies positions where the parallel jaw matches the object's surface, i.e. correlation is high. Symbolic red points are constraints applied to a region (with negative values) that should not collide with the object. Red points physically represent the gripper's palm and wrist.

surface (obtained via a range sensor) and their respective normal (estimated from the point cloud). Our grasp planner's flow diagram is shown in Fig. 1. A grasp plan for a single grasp type is completed in two stages using two data structures: 1) histograms of surface normals ($\mathbf{H_o}$, $\mathbf{H_g}$) and 2) 3D voxel grids ($\mathbf{V_o}$, $\mathbf{V_g}$) computed from point clouds. Subscripts '**o**' and '**g**' denote object and the gripper grasp type respectively. A voxel grid maps sampled points to a discretized 3D grid. A surface normal histogram discretizes surface normals to their respective spherical coordinate angles, inserts them into bins, and shows the surface normal frequency for an object. Grasp contact normals for any grasp type are used to create $\mathbf{H_g}$ and are defined a-priori. The advantage to using surface normal histograms is that they are invariant to translation and rotation from any viewing angle within the world frame. If an object is stationary, the robot's view does not affect the surface normal histogram representation. High frequency noise is filtered because discretizing surface normals smooth these frequencies.

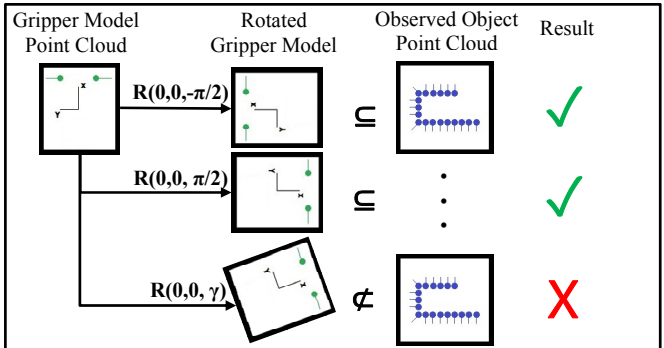

Figure 2. Stage 1 Example for Parallel Jaw to discover Gripper Orientation

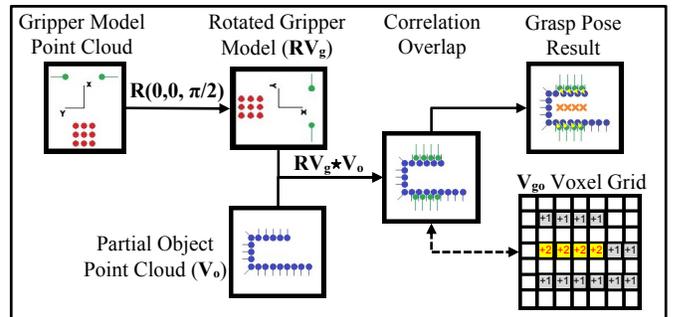

Figure 3. Stage 2 Example for Parallel Jaw to Discover Gripper Orientation

## IV. GRASP PLANNING DETAILS

### a) Stage 1a: Surface Normal Histograms

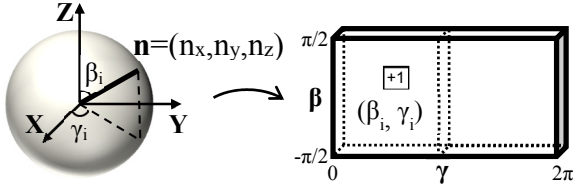

Figure 4. Surface Normal Histogram Data Structure

A surface normal histogram, shown in Fig. 4, is created by representing a point normal, $\mathbf{n}=(n_x,n_y,n_z)^T$ via two spherical angles: 1) elevation / pitch angle $\beta_i$:$[-\pi/2,\pi/2]$ and 2) azimuth / yaw angle $\gamma_i$:$[0,2\pi)$, where 'i' denotes a coordinate $(\beta_i, \gamma_i)$ within a histogram $\mathbf{H}$.

$$\beta_i = \text{acos}(n_z), \gamma_i = \text{atan2}\left(\frac{n_y}{n_x}\right), n_z \neq \pm 1$$

Angles are discretized into uniform bins and incremented. Due to gimble lock when a normal is orientated at north or south poles, a unique solution to $\gamma$ does not exist. For these two cases, all azimuth bins for the polar elevation angle are incremented $\Delta\gamma/2\pi$, where $\Delta\gamma$ denotes bin size. Normals for the gripper represent the grasp type's contact orientation; their orientation faces opposite to an object's surface normals so an inverted normal is inserted into $\mathbf{H_g}$, i.e. $\mathbf{H_g}[-\mathbf{n}]$.

### b) Stage 1b: Matching Histograms

Surface normal histograms are convenient to match shape. All histograms share the same angle resolution. If all non-zero gripper bin locations $\mathbf{H_g}(\beta_i, \gamma_i)$ are also non-zero at the same object histogram bin location $\mathbf{H_o}(\beta_i, \gamma_i)$, there exists a possibility the grasp shape is on the object's surface. However, this is only true for one gripper orientation. To match a grasp type shape for all orientations, $N$ gripper normals $\mathbf{n} \in \mathbb{R}^{3 \times 1}$ are rotated using Euler angles roll $\alpha_g$:$[0,2\pi)$, pitch $\beta_g$:$[-\pi/2,\pi/2]$, and yaw $\gamma_g$:$[0,2\pi)$. A rotation transform is defined as $\mathbf{R} \in \mathbb{R}^{3 \times 3}$. Rotated normals are then mapped to a histogram index, and these indexes are referenced to the object histogram to determine a rank. This method is fast because few gripper contacts are needed to represent grasp type. Note that gripper rotations correspond to bins shifting within $\mathbf{H_g}$.

### c) Stage 1c: Ranking Different Orientations

Gripper orientation ranks are stored in a third surface normal histogram structure, $\mathbf{H_r}$. $\mathbf{H_r}$ bin indexes map gripper orientation, i.e. $\mathbf{H_r}(\alpha_g, \beta_g, \gamma_g)$. A value within a bin contains a rank for each orientation $\mathbf{R}(\alpha_g, \beta_g, \gamma_g)$ the gripper is rotated. Histogram $\mathbf{H_r}$ structure is shown in Fig. 4. For simplicity of visualization, the axis corresponding to roll $\alpha_g$ is not shown, but it is added to $\mathbf{H_r}$ as a third axis to index ranks for all possible rotations $\mathbf{R}(\alpha_g, \beta_g, \gamma_g)$. A bin in $\mathbf{H_r}(\alpha_g, \beta_g, \gamma_g)$ indexes a specific gripper orientation, and the rank value stored quantifies how well $\mathbf{H_g}$ matches with $\mathbf{H_o}$. Ranks for an $N$-contact grasp type are defined as:

$$\mathbf{H_r}(\alpha_g, \beta_g, \gamma_g) = \begin{cases} \sum_{i=1}^{N} \log[\mathbf{H_o}(\beta_i, \gamma_i)], & \mathbf{H_o} \geq \mathbf{R}(\alpha_g, \beta_g, \gamma_g)\mathbf{H_g} \\ & \text{and } \mathbf{H_g}(\beta_i, \gamma_i) \geq 1 \\ 0, & \text{otherwise} \end{cases}$$

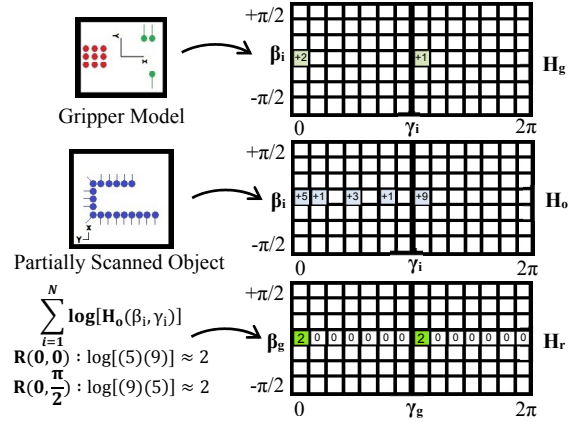

Figure 5. Surface Normal Histogram Matching (Bin Resolution = $\pi/7$)

Rank is a rough measure to indicate the combination of each finger contact choosing a surface normal. If all surface normals match, the logarithmic product is performed at all $\mathbf{H_o}$ bin locations that correspond to non-zero $\mathbf{H_g}$ bin locations. Surface normals do not match when $\mathbf{H_g}(\beta_i, \gamma_i) \geq 1$ and $\mathbf{H_g}(\beta_i,\gamma_i) > \mathbf{H_o}(\beta_i,\gamma_i)$. For this condition, $\mathbf{H_r}(\alpha_g, \beta_g, \gamma_g)$ is set to zero. Practically, thousands of surface normals can represent an object, creating large values within $\mathbf{H_o}$. Rank values can become extremely large using high resolution models. To mitigate this problem and keep ranks values small, a log transformation is performed. Intuitively, rank maximizes when the most surface normals exist for each grasp.

Fig. 5 illustrates a 2D example to create $\mathbf{H_r}(\beta_g, \gamma_g)$ using 2D rotation $\mathbf{R}(\beta_g, \gamma_g)$. All histograms have a resolution set to $\Delta\gamma = \pi/7$ (~25.7°). In this example, only yaw $\mathbf{R}(0, \gamma_g)$ is possible to rotate the gripper. If the gripper elevation angle changes, the same $\mathbf{H_r}$ results would shift up/down along $\beta$-axis for $\mathbf{H_r}$. Fig. 5 demonstrates the logarithmic product rank. By comparing histograms, bins in $\mathbf{H_g}$ align with $\mathbf{H_o}$ when $\gamma_g = \{0, \pi/2\}$. For all other rotations, $\mathbf{H_g}$ bins do not match $\mathbf{H_o}$ and $\mathbf{H_r}(0,\gamma_g) = 0$. Although Stage 1 determines if a gripper orientation matches the grasp type shape to object surface, Stage 2 determines both the scale of the object shape and if it satisfies the gripper's physical constraints.

### d) Stage 2: Voxel Grid to Match Contacts

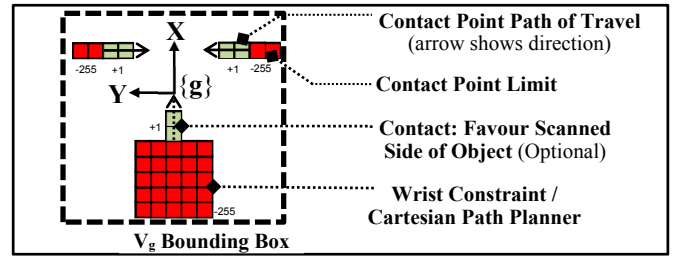

Figure 6. Voxel Grid Model for a 2-Contact Parallel Grip

A collision-free Cartesian plan to grasp an object can be discovered by cross-correlating voxel grid grasp types with partial object voxel grid representations. An object bounding box, $\mathbf{V_o} \in \mathbb{R}^3$, surrounds all object(s) being scanned, and a regular grid discretizes space into voxels. Each voxel is addressed by indexes, $\mathbf{I}=(i,j,k)^T$, where $\mathbf{V_o}(i,j,k)=1$ if an obstacle exists and $\mathbf{V_o}(i,j,k)=0$ if space is empty or unscanned. Shown in Fig. 6, a gripper bounding box, $\mathbf{V_g} \in \mathbb{R}^3$, surrounds the gripper's grasp type being modeled and discretizes space with the same resolution as $V_o$. $V_g(i,j,k)=1$ to model a contact point and its range of motion. This motion is define by a directional vector and length is adjustable. In Fig. 6, two vectors, shown as arrows, point towards each other to model a parallel gripper partially closing. Vector length generalizes the grasp type for different object sizes. A grasp contact assumes its location is graspable for every point along the vector. A long vector length correlates the grasp type with more objects while a short vector length correlates with fewer objects (i.e. the grasp type becomes more size-specific). $V_g(i,j,k)=0$ for empty space. $V_g(i,j,k)=-255$ represents a model's constraints. As stated earlier, points within a point cloud are inserted into a voxel grid. To indicate a point's positive or negative value, its RGB colour value is changed. Specifically, green and red colour channels indicate positive and negative point values respectively.

Although somewhat optional, it is desirable to incorporate additional constraints with the grasp type model; their purpose is to create a desired behaviour for a grasp type. Referring to Fig. 6, the constraints outside the gripper contacts penalize surfaces that are larger than the gripper's "maximum opening", specifies a minimum gap required to place a finger between objects, and helps center the gripper palm towards an object's center. A wrist constraint prevents the gripper's palm from colliding with an object's surface. An additional contact vector from the palm can be added to favour grasp poses along the observed surface of an object. The voxel representations in Fig. 6 and Fig. 8 demonstrate this contact vector; it is added to lateral and tripodal grasp types. Without this vector, cross-correlation can yield valid pose results from both sides of an object (on observed and unobserved sides), and for safety, grasp poses should only exist in observed regions.

Frames {o} and {g} are attached to the center of $V_o$ and $V_g$ bounding box respectively. Object and grasp type voxel grids are built within their respective reference frames; this allows points assigned to $V_g$ to be rotated first and registered to $V_o$ afterwards. The object's point cloud updates $V_o$ after each scan. Cross-correlation between $V_g$ and $V_o$ is performed using the Fast Fourier Transform (FFT). Once voxel grid size and resolution are set, correlation runtime is fixed. Large object point clouds do not negatively impact our algorithm because they are down-sampled to a fixed size voxel grid. In Cartesian space, cross-correlating identical $M$-sized voxel grids $V_g$ with $V_o$ at any gripper orientation $R(\alpha_g,\beta_g,\gamma_g)$ is defined as:

$$V_{go_{(\alpha,\beta,\gamma)}}(x,y,z) = \sum_{i,j,k=0}^{M} R_{(\alpha,\beta,\gamma)} V_g(i,j,k) V_o(x+i, y+j, z+k)$$

Since a voxel grid represents physical space, a grasp type model $V_g$ can be bounded by a relatively small box while the object voxel is adjustable. The largest impact to this algorithm is voxel resolution, but for grasping, resolution can be about as coarse as a gripper's finger width. Correlation also solves two problems with one step: 1) it indicates where a grasp type shape is most similar to the object, and 2) the wrist constraint length determines a collision free Cartesian path for the gripper to move towards an object. A complete gripper pose is found for an $N$-contact grasp type when any voxel $V_{go_{(\alpha,\beta,\gamma)}}(x,y,z)$ is greater than or equal to $N$.

*e)    Stage 2: Verifying Normals and Contacts*

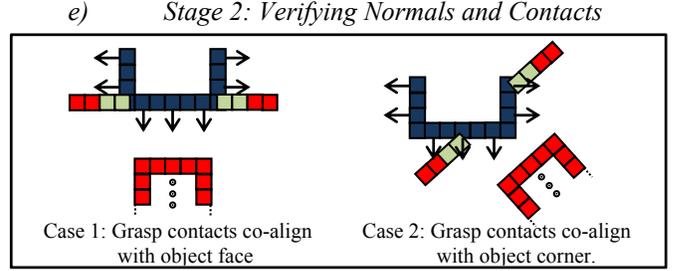

Case 1: Grasp contacts co-align with object face    Case 2: Grasp contacts co-align with object corner.

Figure 7. Removing Illogical Grasps. Contact location (light green) examples for a Partially Scanned Box (blue)

Since shape and scale matching are decoupled and performed in sequential steps, the results after cross-correlation will include locations that do not logically yield a grasp. Two examples are shown in Fig. 7; Case 1 shows a grasp type's contacts creating a plane, or a line in the figure. As a result, the maximum correlation will correspond to a planar surface on one object face instead of two opposing faces. A single face on an object is not graspable. Even if it appears the edges are graspable, edges are not desired because these locations offer the least amount of surface area for the gripper to touch. Case 2 in Fig. 7 shows a similar example, but contacts align at an object's corner (i.e. several faces that are not opposing). If the gripper contacts close around an object's corner, the object will likely slip free.

Verification reasonably checks grasp contacts and their respective normal so that both match the partially scanned object surface. A grasp type position is verified by re-checking the contact normal's direction. At every potential grasp position, a k-d nearest neighbor search is performed for each contact relative to the object(s)[37]. When the closest point on the object is discovered, the inner product of its normal with the gripper normal is taken. The inner product must be greater than a parameterized threshold. Even though this threshold is configurable, we define it as the same size as one surface normal histogram bin, i.e. $\cos(\Delta\gamma/2)$.

V. IMPLEMENTATION DETAILS

All control and motion planning is developed within the Robot Operating System (ROS). Our mobile manipulator comprises of a 3-DOF base (Powerbot), a 6-DOF manipulator (Schunk Power Cube arm), and a 7-DOF 3-fingered Schunk Dextrous Hand (SDH). A Hokuyo URG-04LX planar laser is mounted on the manipulator's wrist as an eye-in-hand sensor and scans all objects in the environment[38]. Its angular resolution is 0.36°. Our robot hardware, software, robot configuration, and approach to grasp an unknown object in an unknown environment is discussed in detail in [38]. The multi-dimensional FFT algorithm used to cross-correlate voxels is developed using the FFTW library[39]. A C++ wrapper is created to integrate FFTW with ROS.

## VI. GRASP MODEL AND PARAMETERS

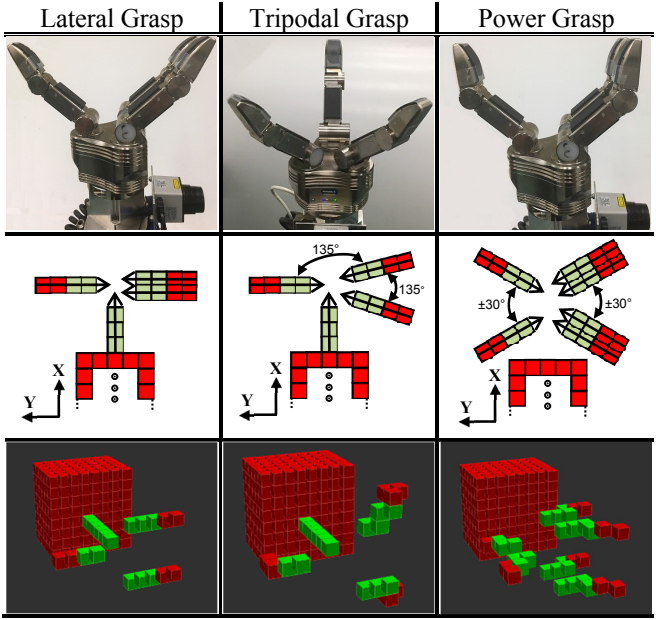

Figure 8. Gripper images for three grasp types (top row),
voxel models for the grasp types (middle row), and
implemented voxel representation (bottom row)

For all experiments, three grasp types (i.e. lateral, tripodal, and power), shown in Fig. 8, are repeatedly searched after scanning a tin can and a cordless drill. The tin can's length and diameter dimensions are 21.0cm x 10.5cm respectively. Ignoring the battery, the cordless drill's length, width, height dimensions are 19.0cm x 6.5cm x 22.0cm. The size difference between objects is important because one grasp type model is discovering a grasp pose for two objects of different widths. A lateral grasp is modelled like a parallel jaw gripper; distal pads (or fingers tips) move towards each other in a pinching motion. A tripodal grasp is similar, but each fingertip is separated by 135° to form a triangular shape that closes. A power grasp is modeled as a box shape that encloses proximal and distal pads around an object at ±30°.

The total contact vector length to model our gripper's motion in voxel grid $V_g$ is 6.0cm, where 4.5cm is applied as a positive value (i.e. $V_g(i,j,k)=0$) and 1.5cm is dedicated as a constraint or negative value (i.e. $V_g(i,j,k)=-255$). These vectors are separated to generalize each grasp type to discover correlations for objects that range between 6.0cm to 12.0cm in diameter. A fourth contact vector from the palm is added to the lateral and tripodal grasp to favour grasp poses along the scanned object's observed side. The power grasp does not have this vector because most of the object needs to be observed before this grasp type is discovered. The wrist constraint is 12.0cm wide (i.e. Schunk SDH width) and extends 10.0cm in length. This length guarantees the gripper can move collision-free 10.0cm along a straight Cartesian trajectory prior to reaching the final grasp pose. Prior to grasping, the gripper moves into the open configuration, similar to configurations shown in Fig. 8. Proximal joint motors engage at a constant velocity to apply a lateral and tripodal grasp; both proximal and distal joint motors engage to complete a power grasp.

The object and gripper voxel size are defined as $V_o$=50x60x30cm and $V_g$=30x30x30cm respectively. The object voxel encapsulate all objects in the world frame. For each grasp type, the gripper wrist rolls (i.e. spins) α:[0,2π), pitches (i.e. pivots up/down) β:[-π/2,π/6], and yaws (i.e. rotates around the object) γ:[0, 2π) at π/6 increments. In total, up to 576 cross-correlations can be performed for each grasp type. In practice, fewer cross-correlations are performed because surface normal histograms $H_o$ and $H_g$ remove poses where the gripper normal do not exist on the object surface. Valid grasp poses are displayed as three different colour arrow markers (i.e. magenta, yellow, and red) for each grasp type (i.e. lateral, tripodal, and power).

## VII. EXPERIMENT & RESULTS

### A. Performance as Voxel Grid Size Increases

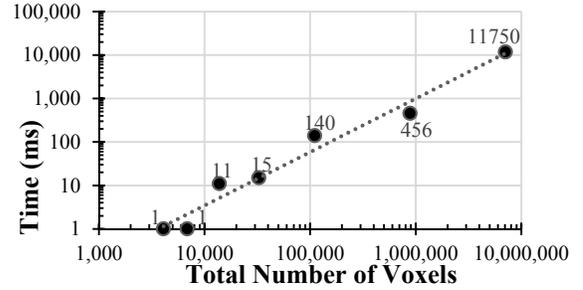

Figure 9. Computation Time Correlating with FFT as Voxel Size Increases

Objects are completely scanned and modeled a-priori with 30,438 points. Point cloud model resolution is expected as less than 3.0mm. The point cloud model is loaded into our algorithm, and time taken to complete the FFT for cross-correlation is measured. This experiment is repeated for different voxel resolutions ($V_{res}$), where $V_{res}$={0.25, 0.5, 1.0, 1.5, 2.0, 2.5, 3.0}cm³. The average FFT computation time and voxel grid size while cross-correlating is recorded and shown in Fig. 9. This figure shows the power relationship between computation time and number of voxels. Feasibly, our system can process grasp results in real-time when voxel grid size is up to 800,000 voxels.

### B. Grasp Results as Voxel Resolution Changes

Fig. 10 visualizes grasp results for voxel resolutions $V_{res}$={0.5, 1.0, 1.5, 2.0, 2.5, 3.0}cm³ from the previous experiment. Magenta, yellow, and red arrows indicate a grasp pose (normal to the gripper palm) for lateral, tripodal, and power grasps respectively. In general, higher resolution reveals more details from the scanned objects and more grasps are discovered. Interestingly, reasonable grasp solutions for all grasp types are consistently found at both low ($V_{res}$=2.0 cm³) and high ($V_{res}$=0.5 cm³) resolutions. For example, a tripodal grasp is available above the tin can to grasp downwards, the drill can be grasped from above, and all lateral/power grasps along the tin can's side face away from the drill to avoid collision. This suggests high resolution may not be ideal and a resolution similar to the width of a finger is reasonable. When $V_{res}$=0.5 cm³, noise (and more details) cause some of these top-down tripodal grasps to be offset by 30°. $V_{res}$={1.0, 1.5}cm³ resolutions smooth noise from the point cloud and clearly select top-down tripodal grasps. On the other hand, low resolution causes a grasp pose's physical location to be rounded by one voxel length; as a result, low resolution introduces a physical position 'offset' error. Our grasp strategy is to trap the object between gripper fingers. A small position error (i.e. < 1.0cm) is likely to be relatively harmless for grasping; however, larger errors cause one finger to bump into the object first, possible tipping the object over or sliding it away, causing a failed grasp attempt.

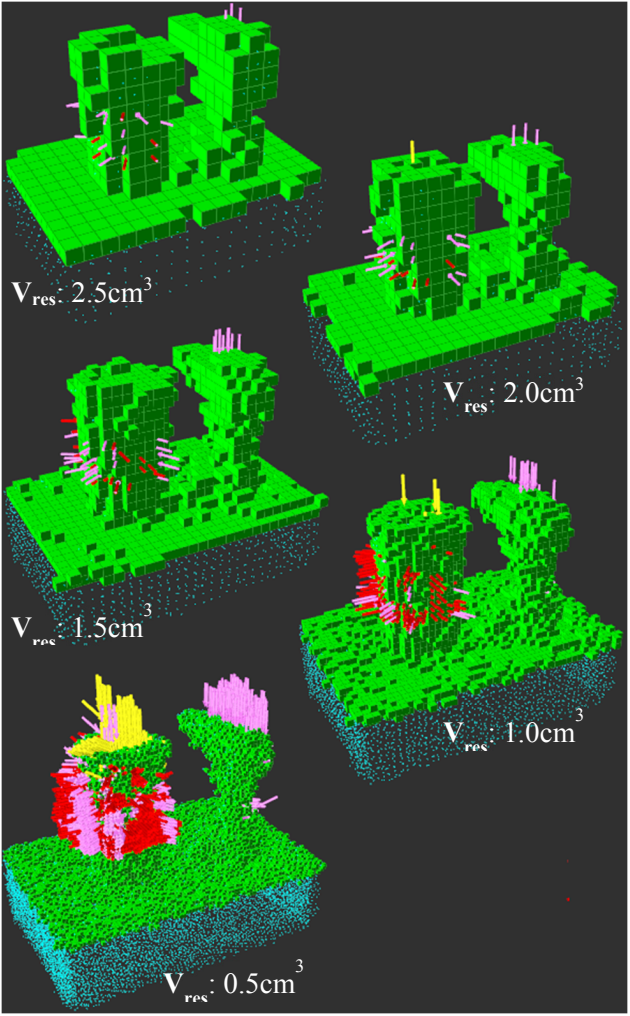

Figure 10. Pose Results at different Voxel Resolutions
Lateral (magenta), Tripodal (yellow), and Power Grasp (red)

## C. Grasp Results Using Incomplete Information

Fig. 11 shows grasp results while scanning an object from five different viewpoints. $V_{res}$=1.5 cm$^3$. Scans are taken counter-clockwise around the objects shown in Fig. 11. Each consecutive scan is registered and merged with the previous scan until a complete object point cloud is created; more details about this process can be found from our previous work[38].

The first scan did not generate any grasp poses. This is expected because a parallel grasp needs two opposing surfaces to be observed to generate a possible result. The second scan in Fig. 11 demonstrates this behaviour as lateral grasps are found top-down and along the tin can's side. All grasp types can be found by the third scan; at this point, the objects' three sides are observed. These results are similar as shown in Fig. 10. In Fig. 11, the point cloud is experiencing 3cm of registration error; this can be observed from the fifth scan, looking at the point cloud's top-right hand corner, where the corners do not align. Please note that our point cloud is down-sampled to the same resolution as $V_{res}$. This error does not significantly affect our algorithm. Pose locations are still centered with respect to each object, and can allow the gripper to trap the object between its fingers.

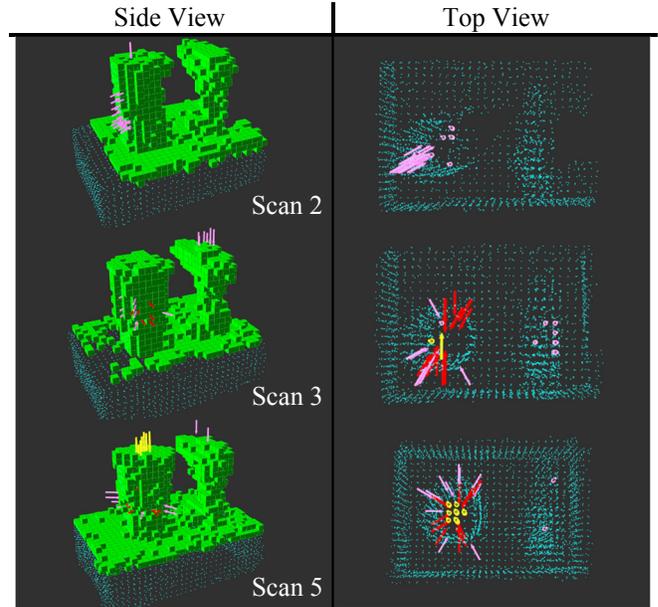

Figure 11. Pose Results while Scanning a Tin Can and Hand Drill
Lateral (magenta), Tripodal (yellow), and Power Grasp (red)

## D. Grasping Autonomously

Our mobile robot platform autonomously scans objects while running our proposed grasping algorithm using $V_{res}$=1.5 cm$^3$. The experiment to autonomously scan and model an unknown object in an unknown region is described in our previous work [38]. Once a grasp pose is identified, the robot randomly selects any grasp pose that satisfies its kinematic constraints, moves toward the final grasp pose, and executes a grasp. Fig. 12a and 12b show the gripper trapping the tin can with a lateral grasp. For a lateral grasp, each grasp location usually has two gripper roll solutions (i.e. α=0° and α=180°). This is why the gripper is rotated differently in these figures. The top view in Fig. 11 and Fig. 12b demonstrate the gripper model's wrist constraint purpose; the wrist constraint identifies grasp pose locations where the Cartesian approach to the object would not collide with neighbouring object.

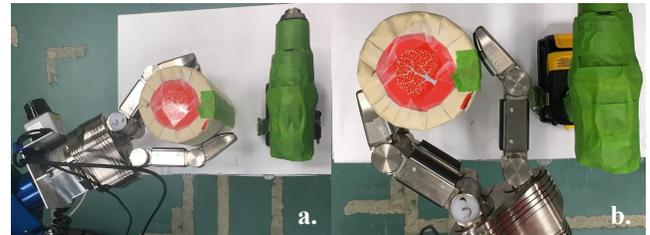

Figure 12. Grasp Examples while Autonomously Scanning Objects

VIII. CONCLUSIONS AND FUTURE WORK

We present a generalized grasping algorithm for objects represented by partial point clouds. We introduce two key ideas: 1) a surface normal histogram can guide a gripper orientation search, and 2) voxel grid representation of a gripper and object can be cross-correlated to discover a grasp pose for any grasp type model. Gripper models for cross-correlation can be generalized to find grasps for objects of different widths and shapes. Experiments show that this algorithm can process grasp poses in near real-time if the voxel count remains below 800,000. Voxel size variation shows grasp results remain consistent for different resolutions.

In future, we plan to show our system grasping an object for different tasks while experiencing uncertainty.